\documentclass[conference]{IEEEtran}
\IEEEoverridecommandlockouts
\usepackage{cite}
\usepackage{amsmath,amssymb,amsfonts}
\usepackage{algorithmic}
\usepackage{graphicx}
\usepackage{textcomp}
\usepackage{xcolor}
\usepackage{multirow}
\usepackage{graphicx}
\usepackage{tikz}
\usepackage{hyperref}
\usetikzlibrary{positioning}

\def\BibTeX{{\rm B\kern-.05em{\sc i\kern-.025em b}\kern-.08em
    T\kern-.1667em\lower.7ex\hbox{E}\kern-.125emX}}
\begin{document}

\title{Enhancing Vehicle Entrance and Parking Management: Deep Learning Solutions for Efficiency and Security\\
\thanks{979-8-3503-1770-1/23/\$31.00 ©2023 IEEE}
}

\author{
    \IEEEauthorblockN{\small Muhammad Umer Ramzan,
    Usman Ali,
    Syed Haider Abbas Naqvi,
    Zeeshan Aslam,
    Tehseen,
    Husnain Ali,
    Muhammad Faheem}
    \IEEEauthorblockA{
       \textit{\small Dept. of Computer Science}, GIFT University, Gujranwala, Pakistan \\
        Emails: \{umer.ramzan, usmanali, 191400106, 191400122, 191400138, 191400118, mfaheem\}@gift.edu.pk \\
    }
}

\maketitle

\begin{abstract}
The auto-management of vehicle entrance and parking in any organization is a complex challenge encompassing record-keeping, efficiency, and security concerns. Manual methods for tracking vehicles and finding parking spaces are slow and wastage of time. In order to solve the problem of auto-management of vehicle entrance and parking, we have utilized the state-of-the-art deep learning models and automate the process of vehicle entrance and parking into any organization.
In order to ensure the security, our system integrated the vehicle detection, license number plate verification and face detection and recognition models to ensure that the person and vehicle are registered with the organization.
We have trained multiple deep learning models for vehicle detection, license number plate detection, face detection and recognition, however YOLOv8n model outperformed from all the other models. Furthermore, License plate recognition, facilitated by Google’s Tesseract-OCR Engine.
By integrating these technologies, the system offers efficient vehicle detection, precise identification, streamlined record-keeping, and optimized parking slot allocation in buildings, thereby enhancing convenience, accuracy, and security. Future research opportunities lie in fine-tuning system performance for a wide range of real-world applications.

\end{abstract}


\begin{tikzpicture}[overlay,remember picture]
\path(current page.north) node(anchor){};
\node[below=of anchor]{2023 25\textsuperscript{th} International Multi Topic Conference (INMIC) };
\end{tikzpicture}

\section{Introduction}
The auto-management of vehicle entry and parking in buildings poses challenges in terms of record keeping, time consumption, and security. Manual recording of vehicle information is time-consuming and makes it difficult for any organization to track vehicle movements effectively. Additionally, finding parking slots can be a time-consuming task. To address these issues, a comprehensive system is needed to detect and identify vehicles, ensuring authorized entry into the building. For this we propose an integrated system that utilize state-of-the-art deep learning techniques to detect vehicle, vehicle's licensed number plate and driver face detection and recognition. Prior to the integration of the system, we have trained multiple deep learning models including Convolutional Neural Network(CNN) on each task of object detection and recognition including vehicle detection and recognition, license number plate detection, and face detection and recognition. The results on each task shows that YOLOv8 (You Only Look Once) model outperforms from all the other deep learning models. The YOLOv8 model is an advanced object detection framework that has gained significant attention in computer vision research in the recent years. It is a deep learning-based architecture designed for real-time object detection tasks. YOLO operates by dividing an image into a grid and predicting bounding boxes and class probabilities within each grid cell\cite{rahman2022densely}. 

Vehicle detection and recognition task from closed-circuit television (CCTV) footage is important while a vehicle reached at the gate of any organization. For this purpose a customized dataset is created and labelled manually to train the model. Furthermore, dataset of Pakistan's licensed number plates are also create and labeled for the task of number plate detection to train the model for number plate detection. Once the number plate of vehicle is detected, the next job is to extract the number from the image of licensed number plate. For this purpose we have utilized the Google’s Tesseract-OCR Engine\cite{smith2007overview} which extracts the characters from the image. This sub system ensure that the vehicle belongs to our organization. However, system should be more secure as someone might have access the organization who does not belongs to that organization as anyone can drive the car that belongs to organization and system verify it and give access by uplifting the barrier. 

In order to solve this security issue, vehicle verification system is combined with driver verification system which will ensure that the vehicle entered into the organization with the registered driver. The driver verification system includes the face detection and recognition, for face detection we have utilized the Haar Cascade classifier\cite{kaur2021face} for the face recognition task as the Haar Cascade classifier is machine learning based approach well-suited for detecting objects with distinct visual features, such as faces or eyes\cite{meena2016approach}. Furthermore, we utilized the DeepFace framework \cite{taigman2014deepface} to find the captured driver image from the employee database of organization. The vehicle will only get access into the organization if the vehicle and the driver belongs to the organization. 

Once the registered vehicle along with the registered driver gets into organization the next step is to park the vehicle at the vacant position which is time consuming task to find the free parking slot to park the vehicle. For this purpose we have developed an android based mobile application which displays the live view of parking area along with free and vacant parking slots. For this purpose, Image processing techniques such as edge detection, thresholding, contour detection, and morphological operations utilized to identify parking slots based on visual cues\cite{crisostomo2019multi}. By analysing the characteristics of the parking  area image, such as colour, texture, and shape, parking slots can be differentiated from the surrounding environment or vehicles. To streamline record-keeping, a NoSQL database called Firebase is employed. This database is flexible, scalable, and offers offline support, ensuring that data remains available even when the device is not connected to the internet.

The manual process of vehicle and driver verification to get access into organization is slow and time consuming task. This paper addresses the issue of manual verification and entrance into organization by automating the process of vehicle detection and verification and driver verification using deep learning techniques and proposed an integrated system that combines various technologies to efficiently manage vehicle detection, identification, driver verification, record-keeping, and vehicle parking automation in any buildings.

\section{Related Work}
The auto-verification of vehicle and parking system consists of different sub system including vehicle detection and verification, number palate detection and recognition, driver's face detection, face recognition and identification of free parking slots in parking area.   There are various methods developed for the implementation of each sub system.
Vehicle detection system is the process of developing an advanced techniques like YOLO v5 to accurately detect, classify, and monitor vehicles in real world traffic scenarios, aiming to enhance safety and contribute to automated driving applications\cite{qazzazcar}. YOLO framework is very famous for vehicle detection and classification system. In complex urban environment real-time object detection is very difficult task, this problem can be solved by training multiple YOLOv5 models with different image sizes and applying ensemble techniques for faster detection in complex urban environments\cite{rahman2022densely}. Deep learning model, specifically YOLO v5 model can be used in intelligent transportation systems\cite{farid2023fast}. Wang et al. developed a lightweight YOLO v5 based network for surveillance video vehicle detection in intelligent transportation systems. By using MobileNetV2 and DSC, they achieved real-time and accurate detection with a 95\% reduction in parameters, providing an efficient solution for highway and urban road safety\cite{wang2022vehicle}. Kaijie Zhang et al. proposed a YOLOv5-based vehicle detection method for accurate and real-time tracking, recognition, and counting. Their algorithm effectively handled challenging conditions such as heavy traffic, night environments, overlapping vehicles, and partial vehicle loss, achieving excellent performance\cite{zhang2022research}. Margret Kasper-Euler’s et al. conducted a study to address rest period planning by detecting heavy goods vehicles at rest areas during winter. They explored the implementation of YOLOv5 with thermal network cameras, focusing on the front and rear features for recognition. The study demonstrated the potential of thermal network imaging for effective vehicle detection in challenging winter conditions, aiming to improve heavy goods
vehicle detection in difficult images with overlaps and cut-offs\cite{kasper2021detecting}.

The Convolutional Neural Networks show promising results for license number plate detection and characters segmentation. There are different types of vehicle license number plate datasets are collected for the purpose of model training.However, majority of the researchers focused on license plate detection and character recognition using datasets like Caltech\cite{griffin2007caltech} along with different techniques such as edge detection, K-means clustering, and CNN for segmentation and recognition\cite{su2017accurate}. 

A huge number of studies has been completed for the task of face detection and recognition just like face detection using OpenCV, exploring algorithms like Adaboost, Haar cascades and Cam Shift\cite{goyal2017face}. OpenCV framework includes a lot of functionalities for the different tasks of machine learning. Researchers have used OpenCV as a main module for features extraction, object detection and recognition\cite{khan2019face}. Apart from OpenCV there is another more accurate face detection algorithm like Haar Cascade outperformed for the task of face detection\cite{kaur2021face}. The task of face detection comes with multiple challenges, such as variations in human body attributes, unclear images, and the need for face tracking in video surveillance. The Viola-Jones method is utilized, where Adaboost is used to train a cascade technique for face detection \cite{meena2016approach}. The paper also introduces terms like facial feature detection, face authentication, expression recognition, and face localization. The face detection process involves capturing an image, reading it using the misread function, and detecting the face within the image. Real-time face recognition system using OpenCV utilizes algorithms like Haar Cascade, Fisherman, and LBP for face detection under varying conditions and recognizes faces based on features and compares them with existing records\cite{khan2019face}, authors provides an approach for face detection and recognition using OpenCV and Python. The paper guides the reader on creating a system using OpenCV and Python, starting with the creation of datasets. The face recognition system performs tasks using Python queries, such as encoding the faces and generating face embeddings. Furthermore The system utilizes the image processing techniques to identify and recognize faces from the given images in the dataset.  

\section{Data Collection and Preprocessing}\label{sec:Data Collection}
Our comprehensive system comprises two main subsystems: vehicle detection and recognition, followed by license plate detection and recognition. The second subsystem involves driver verification, achieved by identifying the driver's face and subsequently comparing it with the employee database. To train the models for each of these subsystems, we gathered a dataset and meticulously labeled it in the subsequent manner:
\subsection{Vehicle Dataset}Our initial task involved collecting a diverse range of images to construct a dataset tailored for the purpose of vehicle detection. To achieve this, we systematically acquired 6000 photographs of vehicle from three different categories, including cars, buses, and trucks. This process was carried out in-person, ensuring that the dataset encompassed a comprehensive representation of these vehicle types. After that we have performed manual labeling by carefully drawing a bounding box around each vehicle and assigning the class name as cars, buses, and trucks according to image. After completing the manual labeling process, we downloaded the labeled data in YOLO format. This format provided a text file for each image, containing the coordinates of the bounding box and class labels of the vehicles. To consolidate the dataset, we combined all the labeled image folders together.
To ensure proper training and evaluation, we split the dataset into three sets called training set, validation set and test set. The training folder contained 60\% of the images, while the validation folder contained 20\% images and the remaining 20\% images are included in test set. 

\subsection{License Number Plate Dataset}
The process of number plate dataset creation is more or less same as we completed the following steps in order to prepare the dataset for number detection. Firstly, we manually captured a total of 600 number plate images including car's and bike's number plate. After that we began the manual labeling process by drawing rectangles around each vehicle's number plate using a drawing tool. We made sure to assign the class name as car or bike number plate based on the image. This labeling process was repeated for all the captured images until each one was properly labeled. The dataset was provided in YOLO format, which included a text file accompanying each image. These text files contained the coordinates and class labels for the detected licence number plate of vehicle. To consolidate the dataset, we combined all the labeled images and their corresponding text files. Once the dataset was complete, we divided the dataset into training, validation and test set for the purpose of fair training. The training set contained 80\% of the images, while the validation set include the 20\% and test set held the 20\% of the images. Importantly, all three datasets retained the necessary labels for each image. 
\begin{figure}
    \centering
    \includegraphics[width=0.48\textwidth]{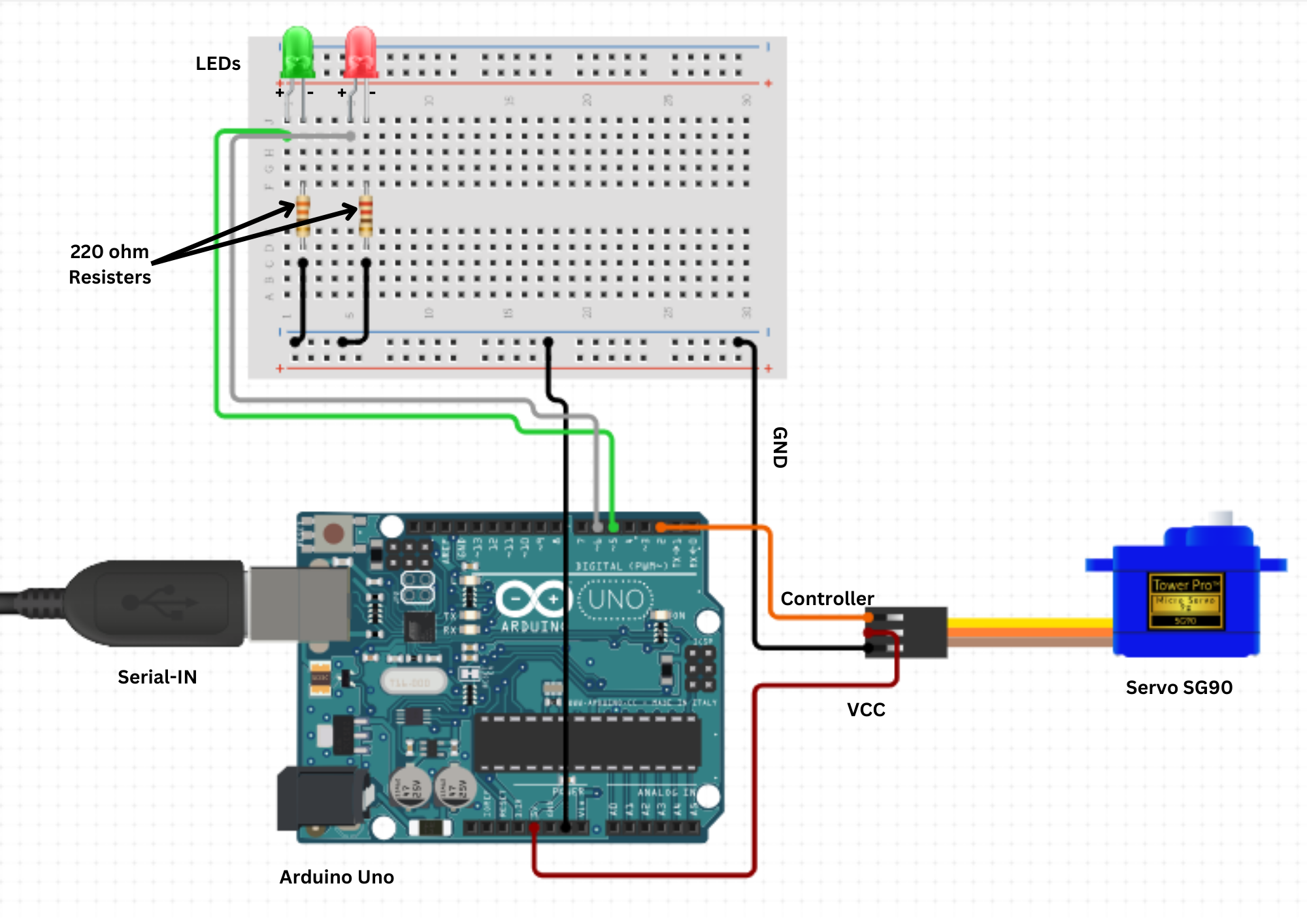}
    \caption{Circuit diagram of automatic barrier control system }
    \label{fig:arduino}
\end{figure}

\section{Methodology}
In this section, we describe the methodology employed to train a Faster R-CNN model, YOLOv5, and YOLOv8 models for simultaneous vehicle detection and number plate detection tasks. Finally we have selected YOLOv8 as it produced best result for vehicle detection and licence number plate detection. The detailed flow of our system is shown in
\begin{figure*}[t]
  \centering
  \includegraphics[width=0.9\textwidth]{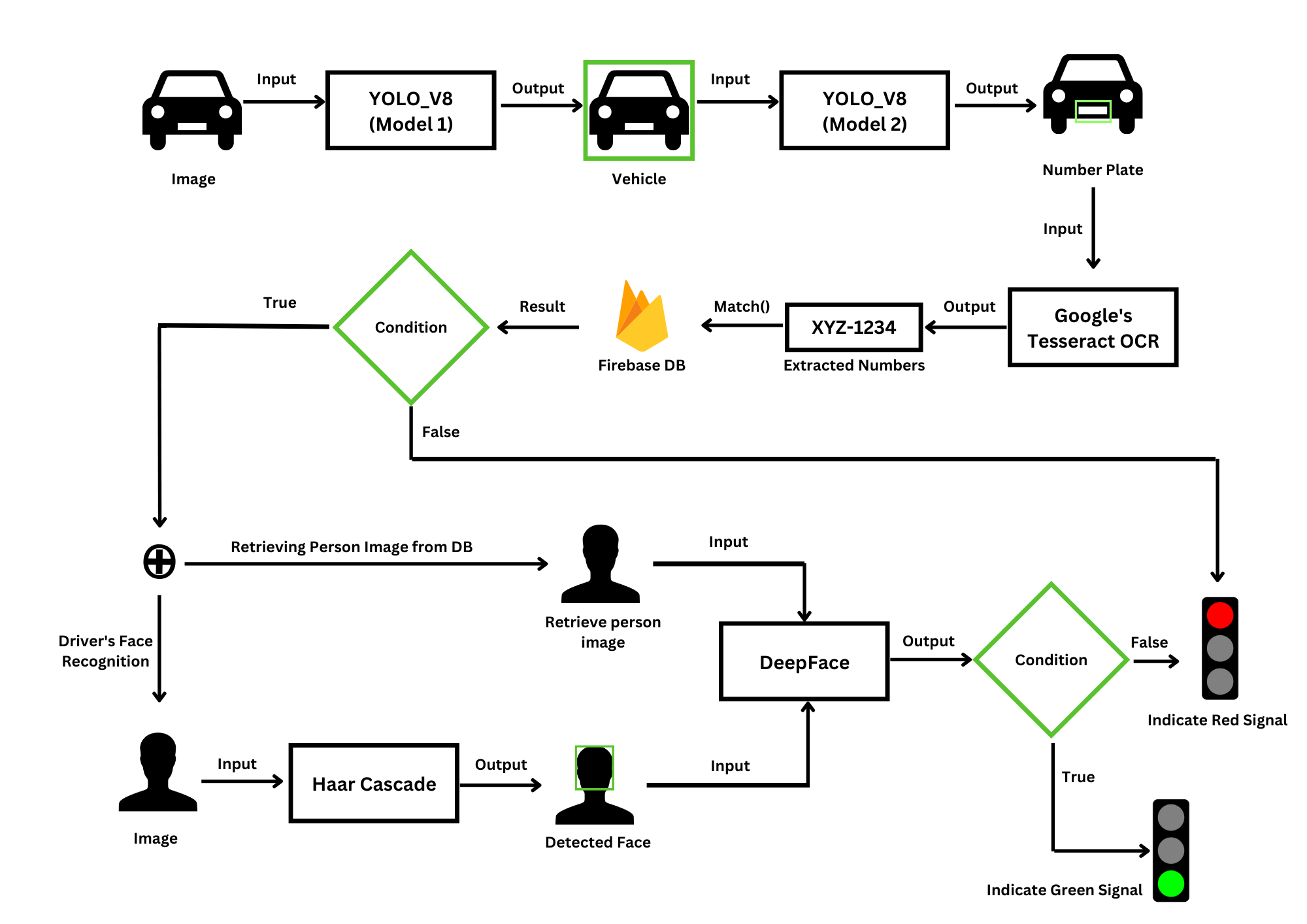}
  \caption{Smart parking system architecture}
  \label{fig:Sys_architecture}
\end{figure*}

Once we have trained the model for vehicle detection and number plate detection, we need to identify that this vehicle belongs to the organization. For this purpose we have utilized Google's Tesseract-OCR engine to extract the licence number from the detected number plate of vehicle using  and match it with the database to authenticate that the vehicle belongs to the organization. 
In order to enhance the security of the system we also include the driver's face recognition and verification system. which will ensure that the driver along with the registered vehicle getting access to the organization. For this purpose we have used Haar cascades that is a machine learning-based approach for face detection. They are widely used for real-time object detection tasks, including face detection.
The key idea behind Haar cascades is to use a set of simple features (Haar-like features) to describe regions of an image and train a classifier to distinguish between object (in this case, face) and non-object regions. The "cascade" in Haar cascades refers to a sequence of stages where each stage contains a set of weak classifiers. Each weak classifier is essentially a simple feature that evaluates whether a region of an image matches a specific pattern.
In order to ensure that the detected image of driver exist in the database of the organization, we employed the DeepFace library fo compare the image of driver with the database. DeepFace is a Python library that provides high-level APIs for deep learning-based face recognition and facial attribute analysis. It uses pre-trained deep learning models to perform tasks like face verification, facial attribute analysis (age, gender, emotion, etc.), and more.
We have also introduced a mobile application which helps the driver to find the free slot in the parking area. We have built a prototype of our system using Arduino micro-controller hardware, the circuit diagram of the system is shown in \ref{fig:arduino}.

\subsection{Data Collection and Preparation}
A diverse dataset consisting of images containing various vehicle types and their corresponding license plates was collected. The dataset was divided into training, validation, and test sets. Each image was represented as $I_i$, where $i$ is the index of the image. The detailed process of dataset preparation explained in section \ref{sec:Data Collection}.

\subsection{Model Architectures and Configuration}

\subsubsection{Faster R-CNN Model}

Faster R-CNN (Region-based Convolutional Neural Network) is a deep learning architecture designed for object detection. It's an improvement over earlier object detection models like R-CNN and Fast R-CNN, offering significantly faster processing speeds while maintaining competitive accuracy.

\subsubsection{YOLOv5 and YOLOv8 Models}

The YOLOv5 and YOLOv8 architectures were chosen for their real-time object detection capabilities. They utilized feature extraction layers and detection layers to predict bounding boxes and class labels. The outputs of YOLOv5 and YOLOv8 models for image $I_i$ are denoted as $O_{\text{YOLOv5}}(I_i)$ and $O_{\text{YOLOv8}}(I_i)$ respectively.

\subsubsection{YOLOv8 Architecture}
The YOLOv8 architecture employs several crucial components for performing object detection tasks. The Backbone comprises a sequence of convolutional layers responsible for extracting pertinent features from the input image. The Spatial Pyramid Pooling (SPPF) layer, followed by subsequent convolution layers, handles feature processing at various scales, while the Upsample layers enhance the feature map's resolution. The C2f module combines high-level features with contextual information to enhance detection accuracy. Lastly, the Detection module employs a combination of convolution and linear layers to map high-dimensional features to the output, which includes bounding boxes and object classes. This overall architecture prioritizes speed and efficiency while maintaining a high level of detection accuracy.

\begin{figure*}[t]
  \centering
  \includegraphics[width=0.9\textwidth]{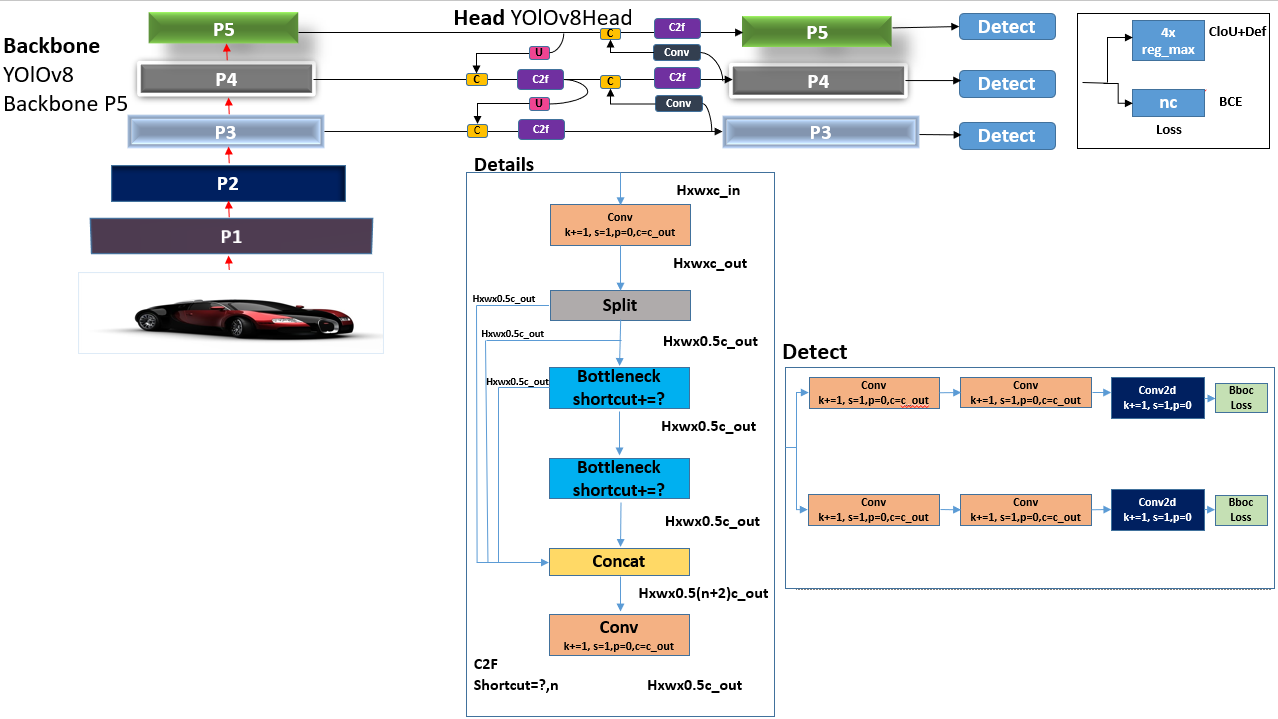}
  \caption{The architectural configuration of YOLOv8 object detection model}
  \label{fig:V8_architecture}
\end{figure*}

In the architecture of YOLOv8 rectangles symbolize different layers, with accompanying labels denoting the layer type (e.g., Conv, Upsample) and any pertinent parameters (such as kernel size or channel count) as shown in \ref{fig:V8_architecture}. The arrows represent the flow of data between these layers, with the direction of the arrow indicating the data's progression from one layer to the next.

\subsection{Loss Functions}
For each model, the training process involved minimizing a combination of localization loss ($L_{\text{loc}}$), objectness loss ($L_{\text{obj}}$), and class prediction loss ($L_{\text{cls}}$). The overall loss function for image $I_i$ is calculated as:

\begin{equation}
L(I_i) = \lambda_{\text{loc}} \cdot L_{\text{loc}}(I_i) + \lambda_{\text{obj}} \cdot L_{\text{obj}}(I_i) + \lambda_{\text{cls}} \cdot L_{\text{cls}}(I_i)
\end{equation}
Where:
\begin{equation}
L_{\text{loc}}(I_i) = \sum_{j} \sum_{\text{boxes}} \mathbb{1}_{ij}^{\text{obj}} \left( (x_j - \hat{x}_j)^2 + (y_j -\hat{y}_j)^2 \right)
\end{equation}
\begin{equation}
    E_{1} =  \sum_{j} \sum_{\text{boxes}} \mathbb{1}_{ij}^{\text{obj}} \left( C_j - \hat{C}_j \right)^2
\end{equation}

\begin{equation}
    E_{2} = \mathbb{1}_{ij}^{\text{noobj}} \left( C_j - \hat{C}_j \right)^2 
\end{equation}

\begin{equation}
L_{\text{obj}}(I_i) = E_{1} + E_{2}
\end{equation}
\begin{equation}
L_{\text{cls}}(I_i) = \sum_{j} \sum_{\text{boxes}} \mathbb{1}_{ij}^{\text{obj}} \sum_{c \in \text{classes}} (p_j(c) - \hat{p}_j(c))^2 
\end{equation}

\begin{align*}
\lambda_{\text{loc}} & \text{ is the weight coefficient for the localization loss} \\
\lambda_{\text{obj}} & \text{ is the weight coefficient for the objectness loss} \\
\lambda_{\text{cls}} & \text{ is the weight coefficient for the class prediction loss}
\end{align*}

The localization loss ($L_{\text{loc}}$) measures the accuracy of predicted bounding box coordinates compared to ground truth coordinates.

The objectness loss ($L_{\text{obj}}$) quantifies the difference between predicted objectness scores and true objectness values, accounting for object and non-object cells.

The class prediction loss ($L_{\text{cls}}$) computes the difference between predicted class probabilities and true class labels.

The $\lambda$ coefficients balance the contributions of each loss component to the overall loss function.

\subsection{Training and Optimization}

The models were trained using stochastic gradient descent (SGD) to minimize the loss function over the training dataset. The weights and biases of the models were updated iteratively using back-propagation. The optimization process aimed to find the optimal model parameters $\theta$ that minimized the loss function:
\begin{equation}
\theta^* = \arg\min_{\theta} \sum_{i} L(I_i)
\end{equation}

\subsection{Performance Evaluation}

The trained models were evaluated using metrics such as Intersection over Union (IOU), mean Average Precision (mAP), and Precision, Recall. These metrics provided insights into the models' accuracy in detecting vehicles and number plates.
\subsubsection{Intersection Over Union IOU}
IOU stands for Intersection over Union. It is a common evaluation metric used in computer vision and object detection tasks to measure the accuracy of object localization. IOU is calculated as the ratio of the area of intersection between the predicted bounding box and the ground truth bounding box to the area of their union.
\[
IOU = \frac{Area\_of\_Intersection}{Area\_of\_Union}
\]
In the context of object detection, a higher IOU score indicates that the predicted bounding box closely aligns with the actual object's location. A perfect overlap results in an IOU score of 1, while no overlap or completely disjoint bounding boxes yield an IOU score of 0.

\subsubsection{Mean Average Precision (mAP)}
To get an overall performance measure for the entire object detection model, you compute the mean of the AP values across all the object classes. This gives you the mAP, which is often reported as a single number representing the model's performance.

\begin{table}[]
\centering
\caption{Vehicle detection results of Faster R-CNN, YOLOv5 and YOLOv8.}
\resizebox{\linewidth}{!}{%
\label{tab:Object detection}
\begin{tabular}{cc|c|c|c|c|}
\cline{3-6}
\multicolumn{1}{l}{}                                & \multicolumn{1}{l|}{} & Precision & Recall & mAP@50 & IOU \\ \hline
\multicolumn{1}{|c|}{\multirow{3}{*}{Faster R-CNN}} & Training Data         & 98.1\%    & 97.8\% & 98.8\% & 96.7\% \\ \cline{2-6} 
\multicolumn{1}{|c|}{}                              & Validation Data       & 96.2\%    & 95.7\% & 98.1\% & 96.1\%  \\ \cline{2-6} 
\multicolumn{1}{|c|}{}                              & Test Data             & 96.4\%    & 96.1\% & 98.2\% & 96.1\%   \\ \hline
\multicolumn{1}{|c|}{\multirow{3}{*}{YOLOv5}}       & Training Data         & 98.3\%    & 98\%   & 99.4\% & 96.9\%     \\ \cline{2-6} 
\multicolumn{1}{|c|}{}                              & Validation Data       & 96.6\%    & 94.8\% & 99\%   & 96.6\%\\ \cline{2-6} 
\multicolumn{1}{|c|}{}                              & Test Data             & 97.8\%    & 98.8\% & 99.1\% &  95\%   \\ \hline
\multicolumn{1}{|c|}{\multirow{3}{*}{YOLOv8}}       & Training Data         & \textbf{99.8\%}    & \textbf{99.4\%} & \textbf{99.5\%} & \textbf{99.2\%}    \\ \cline{2-6} 
\multicolumn{1}{|c|}{}                              & Validation Data       & \textbf{99.7\%}    & \textbf{99.1\%} & \textbf{99.4\%} & \textbf{99.1\%}    \\ \cline{2-6} 
\multicolumn{1}{|c|}{}                              & Test Data             & \textbf{99.2\%}    & \textbf{99\%}   & \textbf{99.4\%} &  \textbf{99.1\%}  \\ \hline
\end{tabular}
}
\end{table}

\begin{table}[]
\centering
\caption{Number detection results of Faster R-CNN, YOLOv5 and YOLOv8.}
\resizebox{\linewidth}{!}{%
\label{tab:plate detection}
\begin{tabular}{cc|c|c|c|c|}
\cline{3-6}
\multicolumn{1}{l}{}                                & \multicolumn{1}{l|}{} & Precision & Recall & mAP@50 & IOU \\ \hline
\multicolumn{1}{|c|}{\multirow{3}{*}{Faster R-CNN}} & Training Data         &  98\%     & 97.1\% & 98.5\%  & 95.1\%    \\ \cline{2-6} 
\multicolumn{1}{|c|}{}                              & Validation Data       &  97.9\%   & 96.2\% & 98.3\%  & 94.2\%    \\ \cline{2-6} 
\multicolumn{1}{|c|}{}                              & Test Data             &  97.6\%   & 96\%   & 98.1\%  & 94.2\%   \\ \hline
\multicolumn{1}{|c|}{\multirow{3}{*}{YOLOv5}}       & Training Data         &  99\%     & 98.9\% & 99.2\%  & 98\%\\ \cline{2-6} 
\multicolumn{1}{|c|}{}                              & Validation Data       &  98.8\%   & 98.9\% & 99.1\%  & 97.9\% \\ \cline{2-6} 
\multicolumn{1}{|c|}{}                              & Test Data             &  98.6\%   & 98.6\% & 99.1\%  & 97.6\% \\ \hline
\multicolumn{1}{|c|}{\multirow{3}{*}{YOLOv8}}       & Training Data         &\textbf{99.5\%}  & \textbf{99.3\%} & \textbf{99.5\%}  & \textbf{98.9\%}  \\ \cline{2-6} 
\multicolumn{1}{|c|}{}                              & Validation Data       & \textbf{99.2\%} &  \textbf{99.1\%} & \textbf{99.3\%}   & \textbf{98.5\%}  \\ \cline{2-6} 
\multicolumn{1}{|c|}{}                              & Test Data             & \textbf{99.1\%}  & \textbf{99.1\%}& \textbf{99.2\%}   & \textbf{98.2\%}  \\ \hline
\end{tabular}
}
\end{table}

\section{Results and Analysis}

Quantitative evaluations were conducted on the validation and test sets to assess the performance of each model. Through a comparative analysis, we aimed to discern the strengths and weaknesses of the Faster R-CNN, YOLOv5, and YOLOv8 models, focusing on detection accuracy and computational efficiency. The results, displayed in the table, underscore that the YOLOv8, a state-of-the-art approach, exhibited outstanding performance across all evaluation metrics in vehicle detection as shown in Table \ref{tab:Object detection}.

Similarly, in the context of license plate detection, a range of deep learning models including Fast R-CNN, YOLOv5, and YOLOv8 were trained. The Table \ref{tab:plate detection} depicting the outcomes verifies that YOLOv8, once again, excelled in terms of all evaluation metrics, establishing its superior performance.

\section{Conclusion}

In order to address the challenge of vehicle entrance and parking management within organizations, we leveraged cutting-edge deep learning models to automate these processes. The existing manual methods, characterized by inefficiency and time wastage, were surpassed by our approach.

Our system's core components, including vehicle detection, license plate verification, and face detection and recognition models, ensure both the security and registration of individuals and vehicles within the organization. Through rigorous training and evaluation, YOLOv8n emerged as the superior model across various tasks.

Furthermore, the integration of Google's Tesseract-OCR Engine for license plate recognition enhances the system's precision and efficiency. As a result, our system provides efficient vehicle detection, accurate identification, streamlined record-keeping, and optimized parking slot allocation within buildings, significantly enhancing convenience, accuracy, and security.

\bibliographystyle{ieeetr}
\bibliography{smartpark}

\end{document}